\RequirePackage{fix-cm}
\documentclass[smallextended]{svjour3}       
\smartqed  
\usepackage{graphicx}
\usepackage{amsmath}
\usepackage{amssymb}
\usepackage{placeins}
\usepackage{subcaption}
\usepackage{caption}
\usepackage{xcolor}
\usepackage{hyperref}
\usepackage{booktabs}
\usepackage{cite}

\newcommand{\revise}[1]{{\textcolor{black}{#1}}}
\newcommand{\revisesecond}[1]{{\textcolor{black}{#1}}}

\newcommand{\model}{GCGRNN}
\newcommand{\eu}{Euclidean}

\begin{document}

\title{Network-wide Multi-step Traffic Volume Prediction using Graph Convolutional Gated Recurrent Neural Network
}


\author{Lei Lin         \and
        Weizi Li        \and
        Lei Zhu*\thanks{*Corresponding author.}
}


\institute{F. Author \at
              first address \\
              Tel.: +123-45-678910\\
              Fax: +123-45-678910\\
              \email{fauthor@example.com}           
           \and
           S. Author \at
              second address
}

\date{Received: date / Accepted: date}

\maketitle

\begin{abstract}
Accurate prediction of network-wide traffic conditions is essential for intelligent transportation systems. In the last decade, machine learning techniques have been widely used for this task, resulting in state-of-the-art performance. We propose a novel deep learning model, Graph Convolutional Gated Recurrent Neural Network (\model{}), to predict network-wide, multi-step traffic volume. \model{} can automatically capture spatial correlations between traffic sensors and temporal dependencies in historical traffic data. We have evaluated our model \revise{using two traffic datasets extracted from 150 sensors in Los Angeles, California, at the time resolutions one hour and 15 minutes, respectively.} The results show that our model outperforms the other five benchmark models in terms of prediction accuracy. For instance, \revise{our model reduces MAE by 25.3\%, RMSE by 29.2\%, and MAPE by 20.2\%, compared to the state-of-the-art Diffusion Convolutional Recurrent Neural Network (DCRNN) model using the hourly dataset. Our model also achieves faster training than DCRNN by up to 52\%.} The data and implementation of \model{} can be found at \url{https://github.com/leilin-research/GCGRNN}. 
\keywords{Deep Learning, Graph Convolutional Gated Recurrent Neural Network, Data-driven Graph Filter, Network-wide Traffic Prediction, Multi-step Traffic Prediction}
\end{abstract}

\section{Introduction}
\label{sec:intro}

Traffic state prediction has been playing a critical role in transportation control and operation in the context of Intelligent Transportation Systems (ITS). Previous studies have focused on short-term traffic state prediction using traditional statistical and machine learning models such as Autoregressive Integrated Moving Average (ARIMA)~\cite{ahmed1979analysis,chandra2009predictions, lin2013short}, Linear Multi-regression~\cite{anacleto2013multivariate}, k-Nearest Neighbor~\cite{zhang2013improved}, Support Vector Machine~\cite{lin2013short}, and Artificial Neural Networks~\cite{lin2018quantifying}. Recently, deep learning~\cite{lecun2015deep} has also been used for short-term traffic state prediction, resulting in state-of-the-art performance~\cite{lin2018predicting,ma2015long,lv2014traffic,polson2017deep}. 

Given significant progress, traffic state prediction still calls for improvement. Spatially, many existing studies predict only the traffic state of \emph{individual} links and intersections measured by a single traffic detector~\cite{chandra2009predictions,lin2013short,anacleto2013multivariate,lin2013k,ma2015long,lin2018quantifying}. Temporally, researchers have focused on \emph{single-step} prediction over various time intervals~\cite{chandra2009predictions,lin2013short,anacleto2013multivariate,lin2013k,ma2015long,wu2016short,polson2017deep,lin2018predicting,cui2019traffic}. However, many essential ITS applications such as vehicle allocation in ridesharing services, travel-route optimization, and staff planning for traffic operation require \emph{multi-step} traffic state prediction over the \emph{entire road network}~\cite{lv2014traffic,cai2016spatiotemporal,Li2017CityEstSparse,lin2018quantifying,luo2019multistep}. 

\begin{figure}
		\centering
		\includegraphics[width=.9\linewidth]{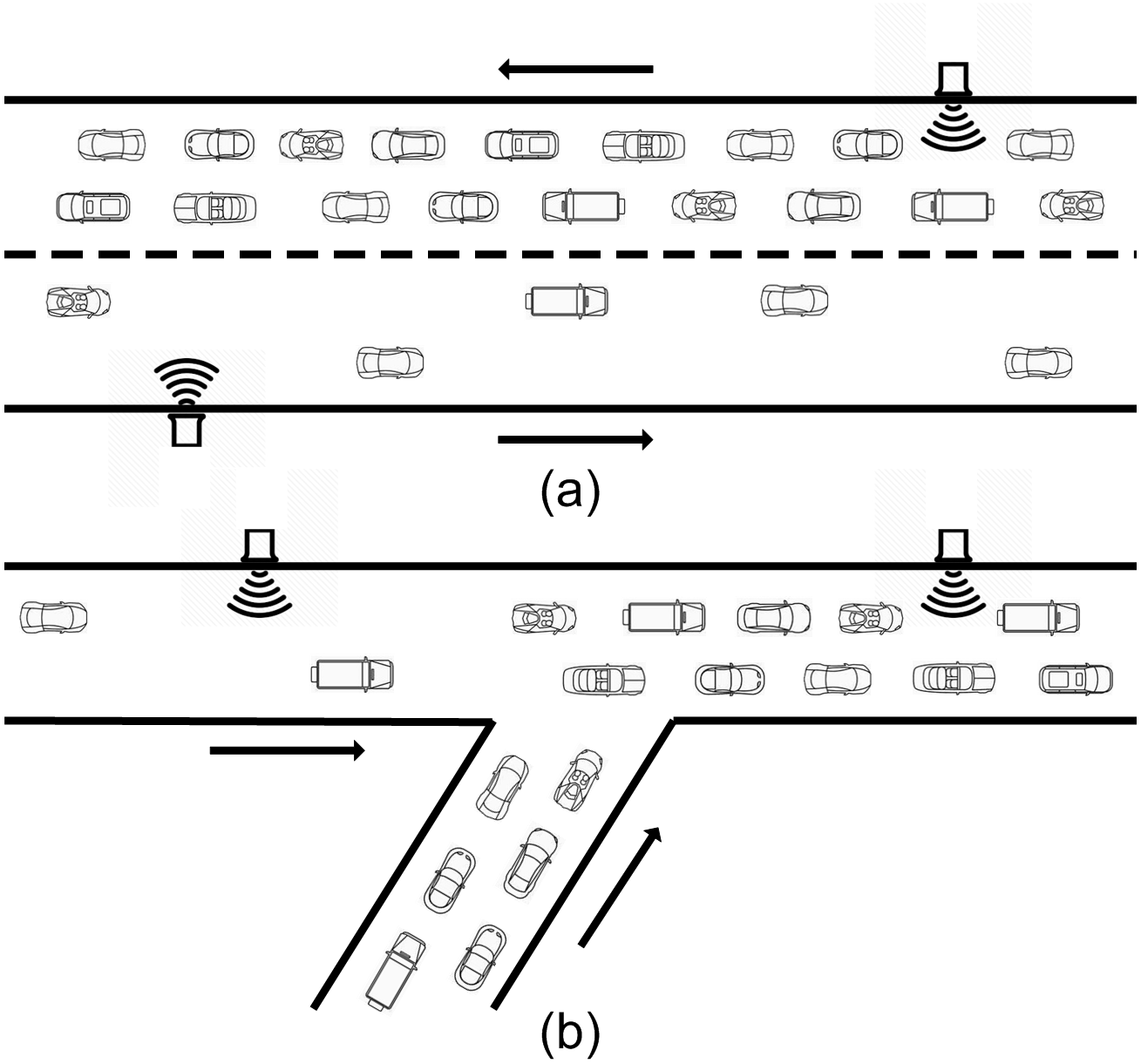}
		\caption{(a) Two spatially-close traffic sensors independently monitor the road conditions. The congestion westbound does not necessarily indicate congestion will develop eastbound. (b) Two topologically-close traffic sensors monitor two locations of one lane. The traffic volume measured by the two sensors could differ substantially due to the in-between on-ramp.}
		\label{fig:sensor}
		\vspace*{-2em}
\end{figure}

One way to improve network-wide traffic state prediction is via exploring spatial-temporal correlations among traffic sensors~\cite{min2011real,lv2014traffic,polson2017deep,li2017diffusion,lin2018predicting}. One common assumption is that measurements from traffic sensors in close proximity are adequate to forecast the traffic states of the links with these sensors. However, this assumption may not hold. For example, as shown in Fig.~\ref{fig:sensor}(a), two spatially-close sensors are independently monitoring traffic flows in two directions. In this case, traffic congestion developed in one direction may not imply traffic congestion in the other direction. To alleviate this problem, researchers have proposed the use of network topology to capture spatial-temporal correlations among traffic sensors~\cite{li2017diffusion,yu2017spatio, cui2019traffic,zhao2019t}. However, this proposal is challenged by the example illustrated in Fig.~\ref{fig:sensor}(b): when an on-ramp locates between two topologically-close sensors, their measured traffic states could vary to a large extent due to the on-ramp. In addition, when an arterial road serves as an alternative to a nearby highway, the traffic states of the arterial road and the highway can be strongly correlated without the two roads being topological neighbors~\cite{polson2017deep}.

Another way to improve network-wide traffic state prediction is to leverage effective models.  Deep learning~\cite{lecun2015deep}, especially Convolutional Neural Networks (CNNs), has been used to improve traffic state prediction. CNNs can automatically extract statistical patterns presented in structured data such as images, but not in structure-varying data such as network traffic data\footnote{\revise{Some studies apply one-dimensional CNN for traffic flow prediction at a fixed site~\cite{du2018hybrid} or to capture spatial features of sensors~\cite{wu2016short}. However, we restrict our discussion to more general uses of CNNs for traffic flow prediction.}}. In order to apply CNNs to network-wide traffic state prediction, one approach is to convert an entire city into a grid map, which is then treated as an ``image''~\cite{ma2017learning}. 
One issue of this approach is the difficulty to choose an effective grid size. A large grid size may fail to satisfy a granularity requirement, while a small grid size will result in high computational costs. Another approach to apply CNNs is through the time-space diagram~\cite{ma2017learning}. The diagram's columns can represent time intervals and the diagram's rows can represent either links or traffic sensors. Each element in the diagram can represent the average speed of a link over a certain time interval. The issue of this approach is that the order of the rows may be arbitrarily chosen and fixed during the learning process. Therefore, the correlations among links cannot be fully captured by the localized filters of CNNs. 

In order to address the limitation of CNNs in processing structure-varying data, researchers have employed Graph Convolutional Neural Networks (GCNNs). GCNNs generalize the convolution operation of CNNs using signal processing and graph theory~\cite{defferrard2016convolutional,sandryhaila2013discrete}. They can naturally handle traffic data that is presented in a graph (e.g., treat traffic sensors as nodes).
As a result, GCNNs have outperformed many traditional statistical and machine learning models, as well as deep learning models~\cite{li2017diffusion,yu2017spatio, cui2019traffic,zhao2019t} in traffic state prediction. Nevertheless, \revise{existing work based on GCNNs largely assumes a strong correlation between two sensors that are spatially or topologically close~\cite{li2017diffusion,yu2017spatio,cui2019traffic,zhao2019t}}. This assumption is limited as illustrated in Fig.~\ref{fig:sensor}. In general, it is challenging to design an effective adjacency matrix---one essential component of GCNNs---to capture correlations among traffic sensors. \revise{A recent study proposes the use of Data-driven Graph Filter (DDGF) with GCNN for predicting network-wide bike-sharing demands~\cite{lin2018predicting}. Their model, GCNN-DDGF, does not need the predefinition of the adjacency matrix.
However, GCNN-DDGF is only tested for single-step bike-sharing demand prediction. }


We propose Graph Convolutional Gated Recurrent
Neural Network (\model{}) for network-wide, multi-step traffic volume prediction. \model{} offers the following advantages:

\begin{itemize}
    \item \revise{embeds a Sequence-to-Sequence Recurrent Neural Network to enable \emph{multi-step} traffic state prediction;}
    \item \revise{can process structure-varying network traffic data and automatically learn the adjacency matrix; and} 
    \item \revise{implements the fast approximation of spectral graph convolution and uses the Gated Recurrent Unit (GRU) to improve training efficiency.}
\end{itemize}

\revise{We have evaluated \model{} using two traffic volume datasets collected from 150 sensors in Los Angeles, California~\cite{Caltrans} at the time resolutions one hour and 15 minutes, respectively. For both datasets, \model{} uses the past 12-step traffic volume to predict the next 12-step traffic volume for all sensors.} The results show that \model{} outperforms two state-of-the-art deep learning models and three traditional statistical models in terms of prediction accuracy. For example, when compared to Diffusion Convolutional Recurrent Neural Network (DCRNN), \model{} reduces MAE by 25.3\%, RMSE by 29.2\%, and MAPE by 20.2\%, using the hourly dataset. \model{} also embraces faster training than DCRNN, achieving speedup by 52\%. More details of our implementations can be found at \url{https://github.com/leilin-research/GCGRNN}.



\section{Related Work}
\label{sec:related}
In this section, we first introduce relevant studies that explore spatial-temporal correlations in short-term traffic state predictions. Then, we discuss models for network-wide, multi-step traffic state predictions. Last, we discuss recent studies that employ GCNNs.

\subsection{Spatial-temporal Correlation}
It has been shown that traffic state data contains spatial-temporal correlations~\cite{lv2014traffic,polson2017deep,min2011real,vlahogianni2007spatio,zou2015hybrid,li2019day}, which can be verified using the cross correlation~\cite{zou2015hybrid} or Hurst exponent~\cite{lin2013short,vlahogianni2007spatio}.
 However, it is challenging to determine spatial-temporal features during the modeling process, for example, identifying the time lag between traffic data from multiple sensors. Some studies apply statistical criteria such as Akaike Information Criterion and Auto-correlation to identify time lags in traffic state data~\cite{zou2015hybrid,li2019day}. Other studies use grid search~\cite{lv2014traffic} or genetic algorithm~\cite{vlahogianni2007spatio} to determine the optimal number of time steps for extracting temporal features. Regarding the relationship of traffic sensors, several studies assume that sensors at the same link~\cite{vlahogianni2007spatio,zou2015hybrid,polson2017deep} or at the adjacent upstream and downstream links~\cite{li2019day} are correlated. However, these assumptions can be limited as  illustrated in Fig.~\ref{fig:sensor}.



\subsection{Network-wide Traffic State Prediction}
One early study of network-wide traffic state prediction is from Min and Wynter~\cite{min2011real}, who apply multivariate spatial-temporal autoregressive model to predict network-wide traffic speed and volume, by taking into account the influence of neighboring links on each link's prediction. 
Cai et al.~\cite{cai2016spatiotemporal} implement an enhanced k-NN model to predict traffic states of 30 links. The nearest neighbors are selected by considering the physical distance, the correlation coefficient, and the connective grade among links. 




In the recent decade, deep learning models have becoming prevalent in network-wide traffic state prediction because of its ability to automatically extract features from raw sensor data. For example, Lv et al.~\cite{lv2014traffic} build a Stacked Autoencoder (SAE) network to capture inherent spatial-temporal correlations embedded in traffic data for prediction.
Other deep learning architectures such as CNNs have emerged in predicting traffic states by converting traffic data into image data~\cite{ma2017learning,wang2019data}. Recent studies have adopted GCNN for network-wide traffic state prediction as it can process structure-varying data natively, we discuss these studies in Sec.~\ref{sec:gcnn}.

\subsection{Multi-step Traffic State Prediction}
In general, three approaches exist for multi-step traffic state prediction. The first approach makes direct multi-step predictions. To provide some examples, Cai et al.~\cite{cai2016spatiotemporal} build an enhanced k-NN model using spatial-temporal state matrices, which outputs 12-step predictions for 30 links. Wang et al.~\cite{wang2019data} adjust the dimension of the CNN output layer to match the number of prediction steps. The second approach builds a single-step prediction model first, then uses consecutive single-step predictions as ``observations'' to make multi-step prediction~\cite{min2011real,zou2015hybrid,yu2016k}. The third approach builds separate prediction models for different prediction horizons. Examples include the models by Lv et al.~\cite{lv2014traffic} and Ma et al.~\cite{ma2017learning}. As an effective architecture for processing sequential data, Seq2Seq-RNN has also been adopted for prediction~\cite{li2017diffusion,Chevalier,liu2019sequence,hao2019sequence,muralidhar2019multivariate}.
Seq2Seq-RNN needs to be trained only once in order to generate multi-step predictions. Similar to the second approach mentioned above, the decoder of Seq2Seq-RNN takes the previous forecasts as ``observations'' for making predictions. 


\vspace*{-2em}
\revise{\subsection{Graph Convolutional Neural Network}\label{sec:gcnn}}



\revise{In order to address the limitation of CNNs in processing structure-varying data, researchers have adopted GCNNs. Based on the convolution operation, GCNN-based studies fall into two categories: spectral-based and spatial-based~\cite{wu2020comprehensive}. Both categories have been applied to network-wide traffic state prediction. Using spectral-based GCNNs, the network-wide traffic data is first converted from the spatial domain to the frequency domain through Fourier transform~\cite{yu2017spatio,zhao2019t,lv_temporal_2020} or wavelet transform~\cite{cui2020learning}. Graph filters are then used in the convolution computation~\cite{kipf2016semi,defferrard2016convolutional}. For spatial-based GCNNs, graph convolutions are conducted directly in the spatial domain. As an example, Diffusion  Convolutional  Recurrent  Neural  Network (DCRNN)~\cite{li2017diffusion} defines the graph convolution as a diffusion process, which is characterized by a random walk and a state transition function based on the adjacency matrix.}

\revise{Previous studies define the adjacency matrix using either \eu{} distance~\cite{yu2017spatio} or network topology~\cite{cui2020learning,lv_temporal_2020, zhao2019t,cui2019traffic,li2017diffusion} between traffic sensors. Some studies build multiple adjacency matrices to capture correlations among nodes. Example adjacency matrices include the Point of Interest matrix which considers the functionality of each node such as residence and business district~\cite{luo2019multistep}, and the similarity matrix which considers historical traffic patterns~\cite{lv_temporal_2020}}. 


\section{Methodology}
\label{sec:methodology}

\revise {In this section, we detail \model{} by explaining its two components: encoder-decoder recurrent neural network and data-driven graph filter.}

\subsection{Encoder-Decoder Recurrent Neural Network}

\begin{figure}
		\centering
		\includegraphics[width=\linewidth]{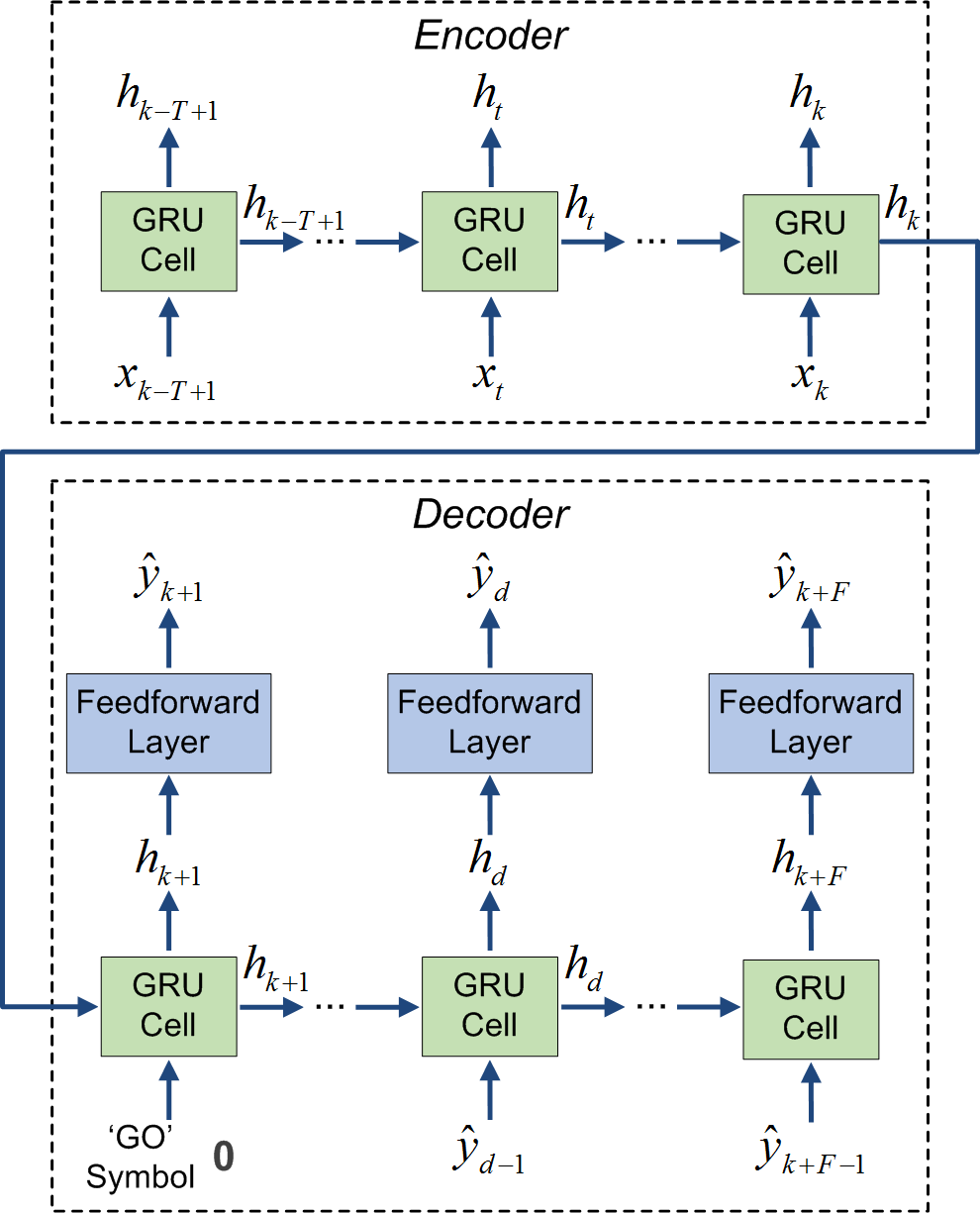}
		\caption{Encoder-Decoder Recurrent Neural Network. For each GRU cell in the encoder, the input is $x_{t=k-T+1,\dots,k}$. The output is the hidden state $h_{d=k+1,\dots,k+F}$. The result the encoder, $h_k$, is passed to the first GRU cell in the decoder to initialize a hidden state. In the decoder, the input to the first GRU cell is a unique `GO' symbol (a zero vector $\mathbf{0}$). Each of the remaining GRU cells in the decoder takes the previous prediction as input. Finally, the decoder outputs $\hat{Y}_k=\{\hat{y}_d\}_{d=k+1}^{k+F}$ as the predictions.}
		\label{fig:architecture}
        \vspace*{-1.4em}
\end{figure}

Assuming a road network has $N$ sensors, the traffic volume data of all sensors in hour $k$ is denoted by $x_k \in \mathbb{R}^{N \times 1}$. The task is to use the historical $T$-step traffic volume data $X_k \in \mathbb{R}^{N \times T},X_k=\{x_t\}_{t=k-T+1}^{k}$ to predict the future $F$-step traffic volume data $Y_k \in \mathbb{R}^{N \times F},Y_k=\{y_d\}_{d=k+1}^{k+F}$. To capture the temporal dependency, \model{} equips encoder-decoder recurrent neural network, which is shown in Fig.~\ref{fig:architecture}. 

Recurrent Neural Network (RNN) is effective in modeling non-linear temporal dependencies in sequential data. Popular RNN cells include Long Short-term Memory (LSTM) cell and Gated Recurrent Unit (GRU) cell. Both types of cells rely on the gate mechanism to modulate the flow of information passing through them. RNNs have been adopted for traffic state prediction~\cite{ma2015long,wu2016short}. In this work, we use GRU cells because they only contain two gates---the update gate and reset gate, resulting in efficient learning~\cite{chung2014empirical}. As shown in Fig.~\ref{fig:architecture}, \revise{the input to a GRU cell at time step $t$ in the encoder includes the historical traffic volume $x_t \in \mathbb{R}^{N \times 1}, {t=k-T+1,\dots,k}$, and the hidden state $h_{t-1} \in \mathbb{R}^{N \times H}$ from the previous time step. We denote by $H$ the number of hidden units in GRU cell. The update gate $z_t \in \mathbb{R}^{N \times H}$ and the reset gate $r_t \in \mathbb{R}^{N \times H}$ inside each GRU cell are calculated as follows:}

\begin{equation} 
    z_t = \sigma\left(\left[h_{t-1},x_t\right]\cdot W_z + b_z\right),
    \label{eq:1}
\end{equation}

\begin{equation}
    r_t = \sigma\left(\left[h_{t-1},x_t\right]\cdot W_r + b_r\right).
    \label{eq:2}
\end{equation}

\noindent The output of the GRU cell is the hidden state $h_t \in \mathbb{R}^{N \times H},t=k-T+1,\dots,k$. In order to compute $h_t$, we first calculate the candidate hidden state $\Tilde{h}_t \in \mathbb{R}^{N \times H}$ using the reset gate defined in Eq.~\ref{eq:2}.

\begin{equation}
    \Tilde{h}_t = tanh\left(\left[r_t \odot h_{t-1},x_t\right]\cdot W_h + b_h\right),
    \label{eq:3}
\end{equation}

\noindent We then compute $h_t$ by combining $h_{t-1}$ and $\Tilde{h}_t$, and using the update gate defined in Eq.~\ref{eq:1}.

\begin{equation}
    h_t = \left(1-z_t\right) \odot h_{t-1}+z_t \odot \Tilde{h}_t.
    \label{eq:4}
\end{equation}

Note that $\sigma$ and $tanh$ represent the sigmoid and \revise{hyperbolic tangent} activation functions, respectively. \revise{The square brackets represent the concatenation of two matrices, e.g., $\left[h_{t-1},x_t\right] \in \mathbb{R}^{N \times \left(H+1\right)}$. The symbol $\odot$ represents pointwise multiplication. $W_z \in \mathbb{R}^{(H+1) \times H}$, $W_r \in \mathbb{R}^{(H+1) \times H}$, and $W_h \in \mathbb{R}^{(H+1) \times H}$ are trainable weights. $b_z \in \mathbb{R}^{1 \times H}$, $b_r \in \mathbb{R}^{1 \times H}$, and $b_h \in \mathbb{R}^{1 \times H}$ are trainable bias parameters.} The reset gate $r_t$ determines which information to keep from the previous hidden state $h_{t-1}$. The update gate $z_t$ decides which information to retain from the previous step and which information to add from the input at the current step. 

In the decoder, $\hat{Y}_k=\{\hat{y}_d\}_{d=k+1}^{k+F}$ represents the predictions. The computing process is following: $h_k$ from the Encoder is fed into the first GRU cell in the decoder to initialize a hidden state. In the decoder, the input to the first GRU cell is a unique `GO' symbol, taking the form of a zero vector $\mathbf{0} \in \mathbb{R}^{N \times 1}$. Each of the remaining GRU cells in the decoder takes previous predictions as input. \revise{The calculation of $h_{d} \in \mathbb{R}^{N \times H}$, $d=k+1, \dots, k+F$, is similar to those described from Eq.~\ref{eq:1}~to~\ref{eq:4}}. $h_d$ is fed into a feedforward layer shown in Fig.~\ref{fig:architecture} to generate the prediction $\hat{y}_d \in \mathbb{R}^{N \times 1}$:
\begin{equation} 
    \hat{y}_d = h_d \cdot W_f,
    \label{eq:5}
\end{equation}

\noindent \revise{where $W_f \in \mathbb{R}^{H \times 1}$ is trainable weights. Note that we did not use activation functions for this feedforward layer and we did not observe any unstable training performance.} Next, we introduce the data-driven graph filter, which can be used to enhance the GRU gate operation of \model{}.

\subsection{Data-driven Graph Filter}
By treating each traffic sensor as a vertex, we represent the whole traffic network as a graph $G=(V,X,\mathcal{E},A)$, where $V$ is a finite set of vertices with size $N$; $X \in \mathbb{R}^{N \times M}$ represents \revise{the sensor data from all the vertices}; $\mathcal{E}$ is a set of edges, $A \in \mathbb{R}^{N \times N}$ is the adjacency matrix, and entry $A_{ij}$ encodes the correlations between two vertices. A normalized graph Laplacian matrix is defined as:

\begin{equation}
    L = I_N - D^{-\frac{1}{2}}AD^{-\frac{1}{2}},
    \label{eq:6}
\end{equation}

\noindent where $I_N$ is the identity matrix and $D \in \mathbb{R}^{N \times N}$ is a diagonal matrix with $D_{ii} = \sum_j A_{ij}$. $L$ is a real symmetric positive-semidefinite matrix which can be diagonalized as:

\begin{equation}
    L = U \Lambda U^{T}, 
    \label{eq:7}
\end{equation}

\noindent where $U=\left[u_0,u_1,\dots,u_{N-1}\right]$ represents orthonormal eigenvectors and $\Lambda= diag([\lambda_0,\lambda_1,\dots,\lambda_{N-1}])$ represents the eigenvalues of $L$. A spectral convolution on the graph is defined as follows:
\vspace*{-0.3em}
\begin{equation}
    g \star X = UgU^TX,
    \label{eq:8}
\end{equation}

\noindent where $g$ is a function of the eigenvalues of $L$. As shown in Eq.~\ref{eq:8}, the whole spectral graph convolution process includes three operations: graph Fourier transform, filtering, and inverse graph Fourier transform.

The computational complexity of Eq.~\ref{eq:8} is $O(N^2)$, which is the result of the multiplication with $U$. Both this computation and the eigen-decomposition of $L$ incur high computational costs. So, we use a simplified spectral convolution on the graph defined as~\cite{kipf2016semi}:
\vspace*{-0.3em}
\begin{equation}
    g_{\Theta} \star X = \Tilde{D}^{-\frac{1}{2}}\Tilde{A}\Tilde{D}^{-\frac{1}{2}}X\Theta,
    \label{eq:9}
\end{equation}

\noindent where $\Theta \in \mathbb{R}^{M \times C}$ is the filter parameter; $\Tilde{A} = A + I_N$ is the summation of the adjacency matrix of the undirected graph $A$ and the identity matrix $I_N$. As a result, $\Tilde{A}$ is the adjacency matrix of an undirected graph in where each vertex connects to itself; $\Tilde{D}_{ii}$ is calculated as $\sum_j \Tilde{A}_{ij}$. With these operations, the computational complexity is reduced to $O(|\mathcal{E}|NC)$. \revise{Eq.~\ref{eq:9} shows that through the spectral graph convolution $g_{\Theta} \star X$, the convolved signal has dimension $N \times C$.}

The spectral graph convolution requires the predefinition of adjacency matrix $A$, \revise{which is nontrivial to determine for network-wide traffic volume prediction, because the adjacency matrix based on either \eu{} distance~\cite{yu2017spatio} or network topology~\cite{li2017diffusion} can be limited (see Fig.~\ref{fig:sensor})}. However, the graph structure is critical to GCNN's performance~\cite{kipf2016semi,defferrard2016convolutional}. To address this issue, Eq.~\ref{eq:8} is redefined to be the following~\cite{lin2018predicting}:
\vspace*{-0.5em}

\begin{equation}
    g_{\hat{A}, \Theta} \star X = \Tilde{D}^{-\frac{1}{2}}\hat{A}\Tilde{D}^{-\frac{1}{2}}X\Theta,
    \label{eq:10}
\end{equation}

\noindent where $\hat{A} \in \mathbb{R}^{N \times N}$ is the Data-driven Graph Filter (DDGF)---a symmetric matrix consisting of trainable filters.

Using Eq.~\ref{eq:10}, now we can integrate the spectral graph convolution into the gate operations in GRU cells. Eq.~\ref{eq:1},~\ref{eq:2},~and~\ref{eq:3} are revised accordingly as follows: 





\vspace*{-0.5em}
\begin{equation}
    z_t = \sigma\left(g_{\hat{A}_z, \Theta_z} \star \left[h_{t-1},x_t\right] + b_z\right),
    \label{eq:11}
\end{equation}

\vspace*{-0.5em}
\begin{equation}
    r_t = \sigma\left(g_{\hat{A}_r, \Theta_r} \star \left[h_{t-1},x_t\right] + b_r\right),
    \label{eq:12}
\end{equation}

\vspace*{-0.5em}
\begin{equation}
    \Tilde{h}_t = tanh\left(g_{\hat{A}_h, \Theta_h} \star \left[r_t \odot h_{t-1},x_t\right] + b_h\right),
    \label{eq:13}
\end{equation}

\noindent \revise{where $\hat{A}_z \in \mathbb{R}^{N \times N}$, $\hat{A}_r \in \mathbb{R}^{N \times N}$, and $\hat{A}_h \in \mathbb{R}^{N \times N}$ are trainable DDGFs; $\Theta_z \in \mathbb{R}^{(H+1) \times H}$, $\Theta_r \in \mathbb{R}^{(H+1) \times H}$, and $\Theta_h \in \mathbb{R}^{(H+1) \times H}$ are trainable weights; and $b_z \in \mathbb{R}^{1 \times H}$, $b_r \in \mathbb{R}^{1 \times H}$, $b_h \in \mathbb{R}^{1 \times H}$ are trainable bias 
parameters.}

\section{Experiments}
\label{sec:exp}

\subsection{Datasets}
\label{sec:data}

\begin{figure}
		\centering
		\includegraphics[width=\linewidth]{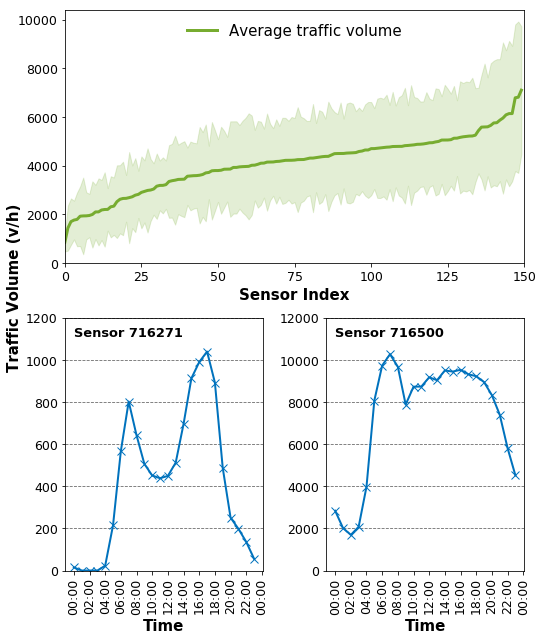}
		\caption{TOP: average traffic volume from 150 sensors in our hourly dataset. \revise{BOTTOM-LEFT: hourly traffic volume of a single lane on Feb. 1, 2018, recorded by Sensor 716271 on a local road. BOTTOM-RIGHT: hourly traffic volume of six lanes on Feb. 1, 2018, recorded by Sensor 716500 on a highway}.}
		\label{fig:data}
		\vspace*{-1.4em}
\end{figure}

\revise{We use two traffic volume datasets for our experiments at the time resolutions one hour and 15 minutes, respectively. The choice of these two time resolutions is to conform with and provide comparisons to previous studies (e.g., one hour~\cite{tan2009aggregation,dunne2013weather,lin2014line} and 15 minutes~\cite{aslam2012city}).} The hourly dataset is collected from 150 sensors in Los Angeles, California, from Jan 1, 2018, to Jun 30, 2019~\cite{Caltrans}, which contains 13~104 hours of traffic volume per each sensor. The mean traffic volumes of all sensors are sorted from low to high and shown in Fig.~\ref{fig:data}~TOP. As examples, the 24-hour traffic volume of a weekday of Sensor 716271 and Sensor 716500 (randomly selected) are shown in Fig.~\ref{fig:data}~BOTTOM. \revise{Sensor 716271, which records the traffic volume of a single lane}, has two peak hours with the larger one showing about 1~000 vehicles appearing at 17:00. Sensor 716500, \revise{which locates on highway and monitors the traffic volume of six lanes}, shows more than 10~000 vehicles appearing in the morning, and more than 8~000 vehicles appearing in the afternoon. \revise{The 15-minute dataset is from the same 150 sensors from Jan 1, 2019 to Jun. 30, 2019, which contains 17~376 data entries per sensor.}

\begin{figure}
		\centering
		\includegraphics[width=\linewidth]{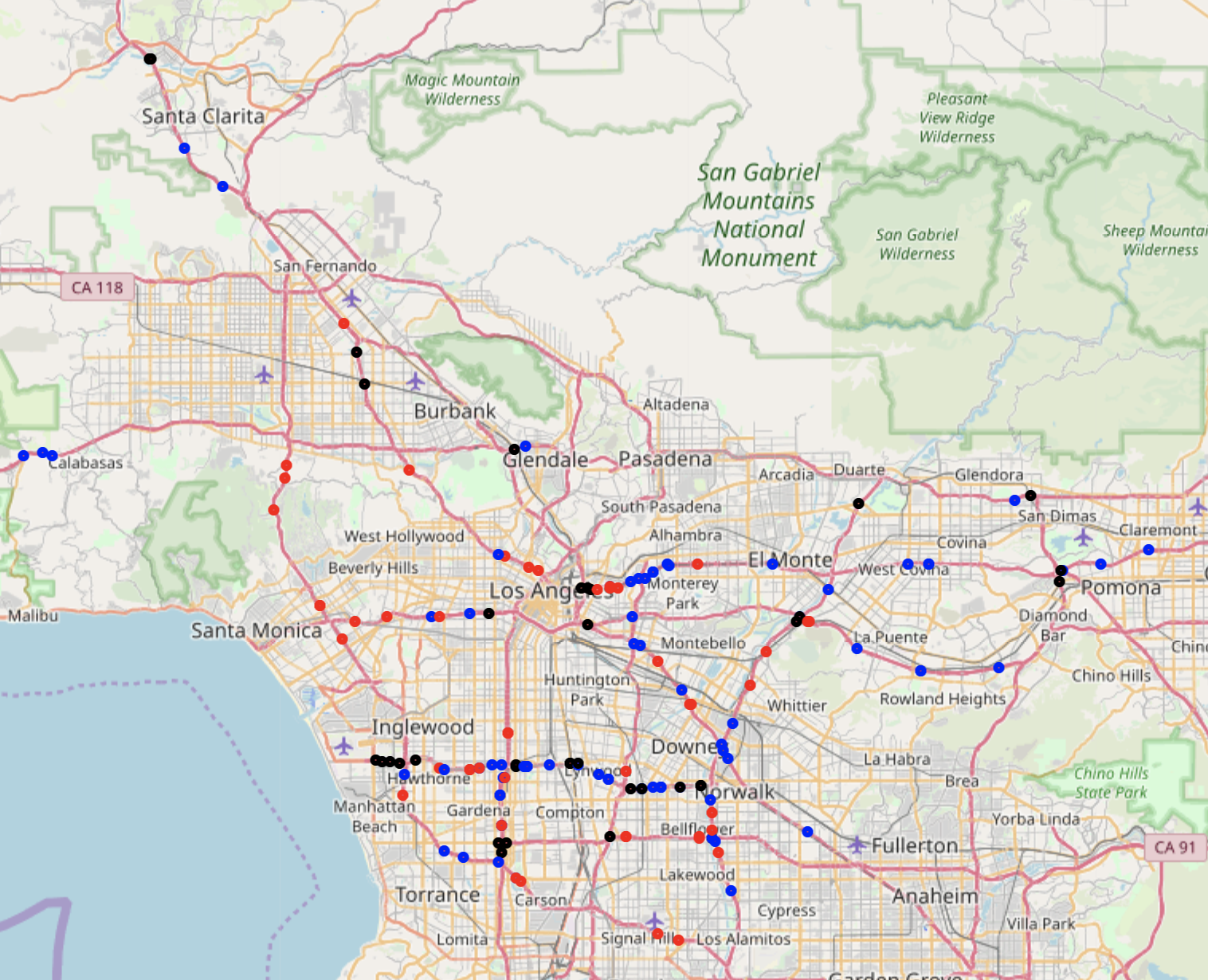}
		\caption{Locations of the 150 Sensors in our datasets. Sensors with the average hourly traffic volume in the top 25 percentiles are red; sensors with the average hourly traffic volume in the bottom 25 percentiles are black. The rest are blue. The sensors are mainly located along highways. Sensors located at the west side of the network show higher traffic volumes than the sensors located at the east side of the network.}
		\label{fig:map}
		\vspace*{-1.4em}
\end{figure}

Fig.~\ref{fig:map} shows the locations of all 150 sensors in our datasets. Red and black points represent the sensors with the mean hourly traffic volumes in the top and bottom 25 percentiles, respectively. The blue points represent the rest of the sensors. Most sensors are distributed along highways. We also observe that the west side of the network has more sensors denoted in red which have relatively higher traffic volume than the sensors located on the east side of the network. 

Our goal is to predict 12-step traffic volume for all sensors using their historical 12-step traffic volume. \revise{We prepare training, validation, and test sets with the ratio 7:1:2 using a moving window approach. For example, the sample at step $k$ takes the form of $\left(X_k, Y_k\right)$, where $X_k=\left[x_{k-T+1},\dots,x_k\right] \in \mathbb{R}^{N \times T}$ represents historical traffic volumes, and $Y_k=\left[y_{k+1},\dots,y_{k+F}\right] \in \mathbb{R}^{N \times F}$ represents the predicted traffic volumes. The step size of the moving window is one. So, for the hourly traffic volume dataset, the data of any 12 hours of a day can serve as our input---leading to 9~157 training samples, 1~308 validation samples, and 2~616 test samples. All three sets are normalized using Z-score normalization based on the mean and the standard deviation of the training set. For the 15-minute dataset, we have 12~148 training samples, 1~735 validation samples, and 3~471 test samples. } \revisesecond{The moving window approach is a commonly used  approach for traffic state prediction~\cite{lv2014traffic,ma2017learning,cui2019traffic,vlahogianni2008temporal}. It can better capture temporal dependencies embedded in the data and generate enough training samples to prevent model overfitting, thus enhancing the prediction accuracy and model generalisability.}

\subsection{Models}

 We contrast our work to several benchmark models including statistical models and deep learning models, which we briefly introduce in the following:

\begin{itemize}
    \item Diffusion Convolution Recurrent Neural Network (DCRNN). DCRNN outperforms multiple state-of-the-art machine learning models for network-wide, multi-step traffic speed prediction~\cite{li2017diffusion}. DCRNN requires a pre-defined adjacency matrix $A$ built with road network features, which we compute as follows:
    \begin{equation} 
    A_{ij} = \exp{\frac{-dist(v_i,v_j)}{\sigma^2}},
    \end{equation}
    \noindent where $dist(v_i,v_j)$ is the spatial distance between the sensors $v_i$ and $v_j$; $\sigma$ is the standard deviation of the distance. In order to keep the sparsity of the adjacency matrix, $A_{ij}$ is set to 0 if $A_{ij} \le 0.1$.
    \item Sequence-to-Sequence Recurrent Neural Network (Seq2Seq-RNN). RNN is a class of neural networks designed for learning sequential data, where outputs from previous steps are fed into the current step as inputs. 
    \item Vector Autoregression (VAR). Autoregression (AR) has been widely used in traffic volume prediction~\cite{lin2013short,lin2014line}. VAR has adapted AR by taking high-dimensional time series as input for predicting multi-sensor traffic volumes.
    \item Linear Regression (LR). LR is implemented for each sensor to predict 12-hour traffic volume using the corresponding historical 12-hour traffic data.
    \item Historical Average (HA). HA is implemented by taking the mean of the traffic volume over the same time span across different days as input.
\end{itemize}

\subsection{Evaluation Metrics}
Three metrics are used in evaluation, namely mean absolute error (MAE), mean absolute percentage error (MAPE), and root mean square error (RMSE). They are computed as follows:
\vspace*{-1em}
\begin{equation} 
    MAE = \frac{1}{N_{test}*F*N}\sum_{i}^{N_{test}}\sum_{d}^{F}\sum_{n}^{N}|y^{i}_{dn}-\hat{y}^{i}_{dn}|,
\end{equation}

\begin{equation} 
    MAPE = \frac{1}{N_{test}*F*N}\sum_{i}^{N_{test}}\sum_{d}^{F}\sum_{n}^{N}\frac{|y^{i}_{dn}-\hat{y}^{i}_{dn}|}{y^{i}_{dn}},
\end{equation}

\begin{equation} 
    RMSE = \sqrt{\frac{1}{N_{test}*F*N}\sum_{i}^{N_{test}}\sum_{d}^{F}\sum_{n}^{N}(y^{i}_{dn}-\hat{y}^{i}_{dn})^2},
\end{equation}

\noindent where $N_{test}$ is the number of test samples, $F=12$ is the prediction horizon, and $N=150$ indicates the number of sensors. $y^{i}_{dn}$ and $\hat{y}^{i}_{dn}$ are the actual and predicted traffic volumes at the $d$th prediction step for the $n$th sensor and the $i$th sample in the test set, respectively. \revise{In this work, we exclude ground-truth traffic volumes that are equal to 0 in the calculation of all three metrics. The percentage of these data entries is $0.008\%$ in the test set from the hourly dataset and $0.03\%$ from the 15-minute dataset.}

\subsection{Model Training}
The training of deep learning models such as \model{}, DCRNN, and Seq2Seq-RNN is not trivial. In this study, we tune hyperparameters via grid search. \revise{The training objective is to minimize MAE. GRU cell is chosen to be part of RNN for all three deep learning models. We use the Adam optimizer~\cite{kingma2014adam} with learning rate 0.01. The learning rate will decay to 0.001 after the first 20 epochs. The number of epochs is set to 300 for \model{} and DCRNN, and 100 for Seq2Seq-RNN. The training batch size is set as 32 for all three deep learning models.} For all three deep learning models, an early-stopping mechanism is adopted to avoid overfitting: if the MAE of the validation set is not reduced for 50 consecutive epochs, we stop the training. The test set is used on the model that has the lowest MAE on the validation set. \revise{All implementations are done using Tensorflow 1.9.0.}. 


\revise{We find that \model{} can be trained much faster than DCRNN. Using the hourly dataset,} the training time of \model{} is about 18.1 seconds per epoch, of DCRNN is about 37.8 seconds per epoch. \revise{Using the 15-minute dataset, the training time of \model{} is about 22.9 seconds per epoch, of DCRNN is about 48.2 seconds per epoch. Both cases show approximately 52\% reduction in training time.} We conduct all experiments using an Intel(R) Core(TM) i7-6820HK CPU, an Nvidia GTX 1080 GPU, and 64G RAM. The training efficiency of \model{} is because we use a simplified spectral graph convolution~\cite{kipf2016semi} (instead of using the diffusion convolution). 


\begin{table}[ht] 
    \centering
    \scalebox{1}{
    \begin{tabular}{cccc}
        \toprule
        & MAE (v/h) & RMSE (v/h) & MAPE  \\        
        \cmidrule(l){2-4}   
        Model & 1 hr / 15 mins & 1 hr / 15 mins & 1 hr / 15 mins  \\ 
        \midrule
        \textbf{\model{}} & \textbf{343.4 / 312.3} & \textbf{540.7 / 493.3} & \textbf{14.2\% / 11.1\%} \\
        \midrule
        DCRNN & 459.8 /489.3  & 763.8 / 794.1 & 17.8\% / 16.7\%
        \\
        \midrule
        Seq2Seq-RNN & 521 / 522.9 & 821.5 / 832.9 & 21.1\% / 18.6\%
        \\
        \midrule
        VAR & 636.3 / 528.2 & 890.2 / 753.8  & 24.3\% / 20.2\%
        \\
        \midrule
        LR & 788.1 / 567.2 & 1087.4 / 852.5 & 33.4\% / 20.4\%
        \\
        \midrule
        HA & 753 / 593.1 & 1086.8 / 890.5 & 35.4\% / 23.6\%
        \\
        \bottomrule
    \end{tabular}}
    \caption{Evaluation results. \model{} achieves the best performance across all experiments using both datasets. All three deep learning models outperform the traditional statistical models VAR, LR, and HA, \revise{except for DCRNN and Seq2Seq-RNN, whose RMSE is lower than that of VAR using the 15-minute dataset.}}
    \label{tb:result}
	\vspace*{-1.4em}
\end{table}

\subsection{Overall Prediction Performance} 
The evaluation results using the two datasets are summarized in Table~\ref{tb:result}. \model{} has the best overall performance. \revise{Using the hourly dataset}, deep learning models \model{}, DCRNN, and Seq2Seq-RNN outperform traditional models VAR, LR, and HA in all measures. \revise{The performance of the traditional models using the 15-minute dataset is better than using the 1-hour dataset. Although deep learning models still achieve higher accuracy in most cases, the RMSE of VAR is even slightly lower than that of DCRNN. This may be because the 12-step 15-minute traffic volumes fluctuate less than the 12-step hourly traffic volumes, thus easier for traditional models making predictions.}   Contrasted to the Seq2Seq-RNN model, the graph convolution operation in \model{} and DCRNN is the main reason for the performance increase. \revisesecond{Note that although our model outperforms all the other models, the prediction accuracy may be insufficient for certain traffic management tasks such as high-precision traffic assignment.}


\begin{figure*}[ht]
		\centering
		\includegraphics[width=\textwidth]{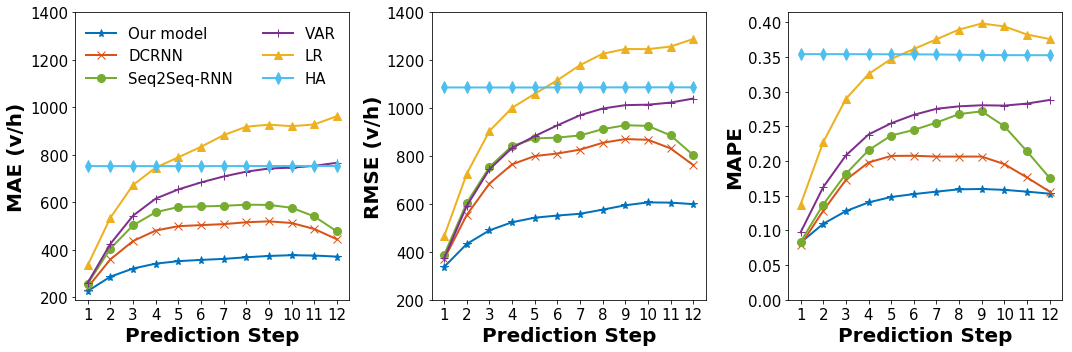}
		\caption{Temporal prediction performance of six models using \revise{the hourly dataset}. \model{} outperforms the other models for all 12 steps in all measures, except for the first step, where it is slightly worse than DCRNN in MAPE.} 
		\label{fig:temporal}
\end{figure*}

\begin{figure*}[ht]
		\centering
		\includegraphics[width=\textwidth]{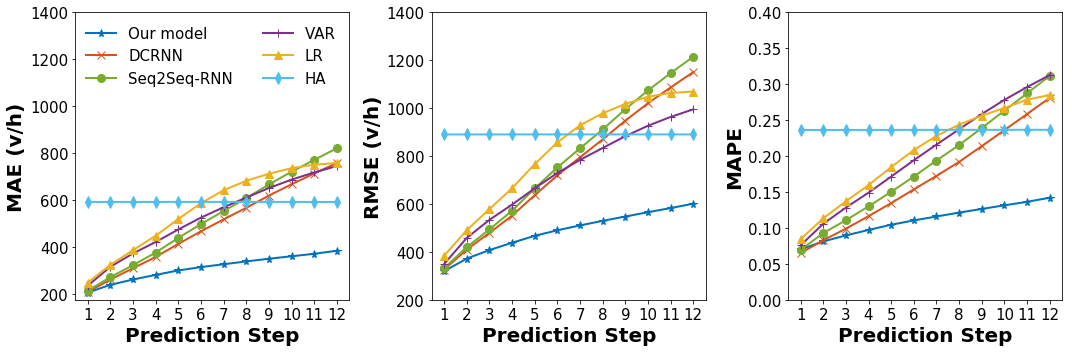}
		\caption{\revise{Temporal prediction performance of six models using \revise{the 15-minute dataset}. \model{} outperforms the other models for all 12 steps in all measures, except for the first step, where it is slightly worse than DCRNN and Seq2Seq-RNN in MAPE. 
		}} 
		\label{fig:15mins}
		\vspace*{-2em}
\end{figure*}

\begin{figure*}
		\centering
		\includegraphics[width=\textwidth,height=.23\textheight]{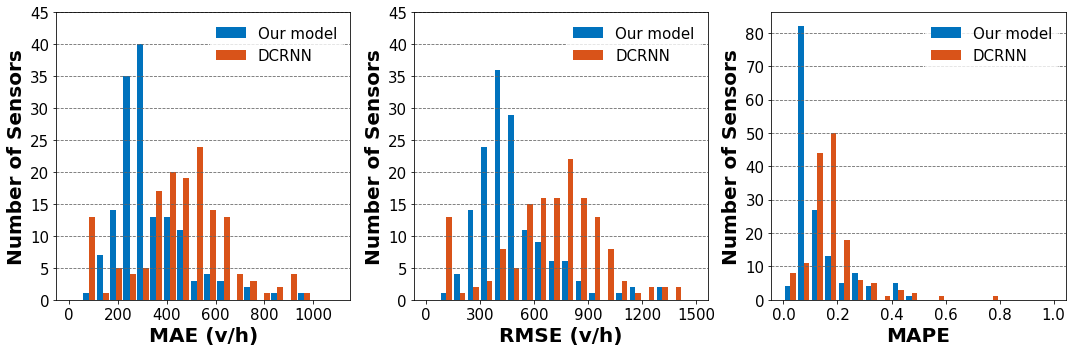}
		\caption{Performance comparison of \model{} and DCRNN using MAE, MAPE, and RMSE on all 150 sensors. \model{} has smaller errors than DCRNN across all measurements. \revisesecond{Note that although the overall MAPE for 12-step predictions of 150 sensors is around 14\%, 4 sensors have their MAPE under 5\% and 82 sensors have their MAPE under 10\%.}}
		\label{fig:close}
\end{figure*}

\begin{figure}
		\centering
		\includegraphics[width=\linewidth]{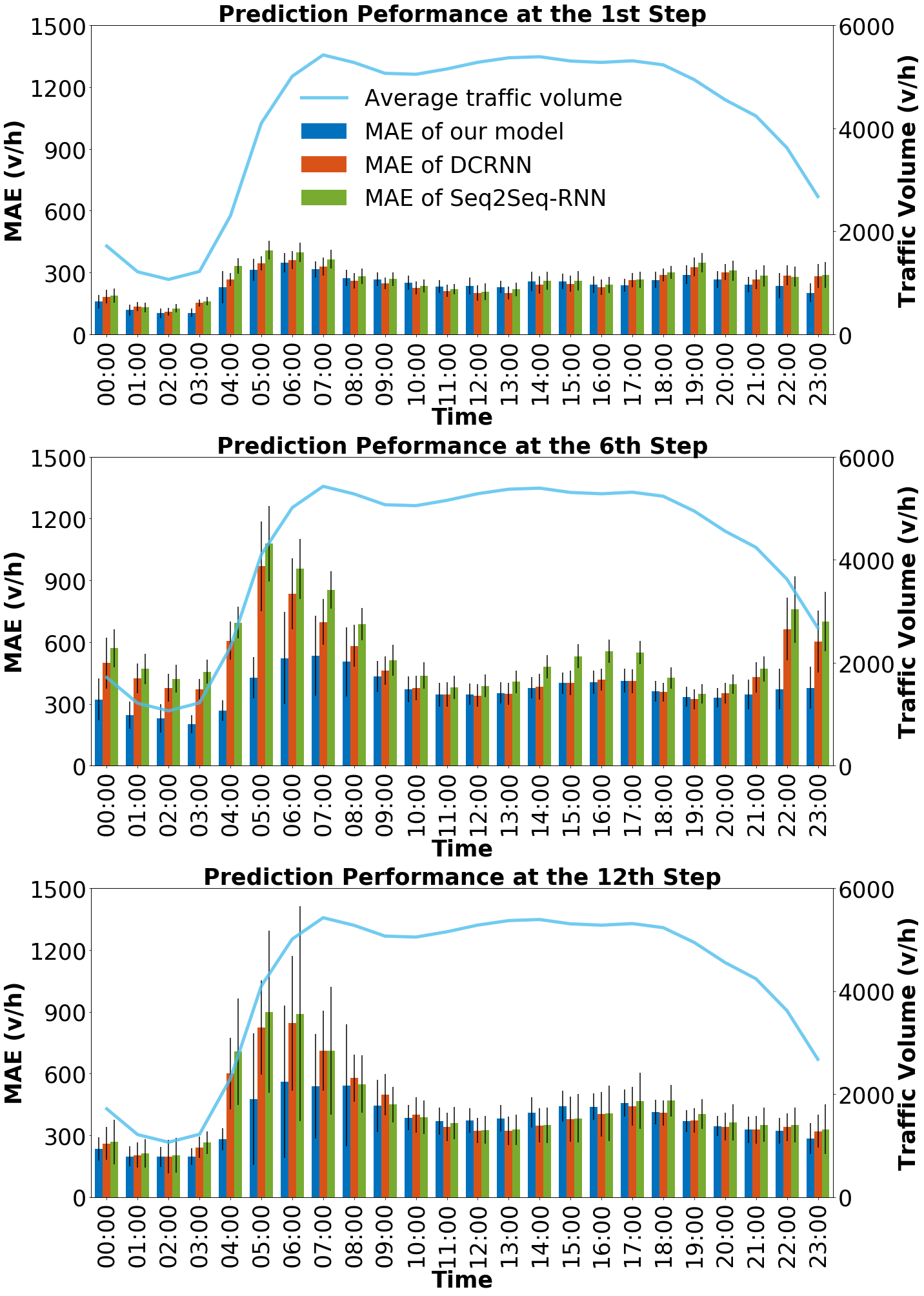}
		\caption{\revise{Predictions at the 1st, 6th, and 12th prediction steps of our model, DCRNN, and Seq2Seq-RNN across 24 hours on weekdays. All three models perform similarly across all hours at the 1st step. Our model outperforms DCRNN and Seq2Seq-RNN over the early-morning hours (04:00-08:00) at the 6th and 12th prediction steps.}} 
		\label{fig:dynamics}
\end{figure}

\begin{figure}
		\centering
		\includegraphics[width=\linewidth]{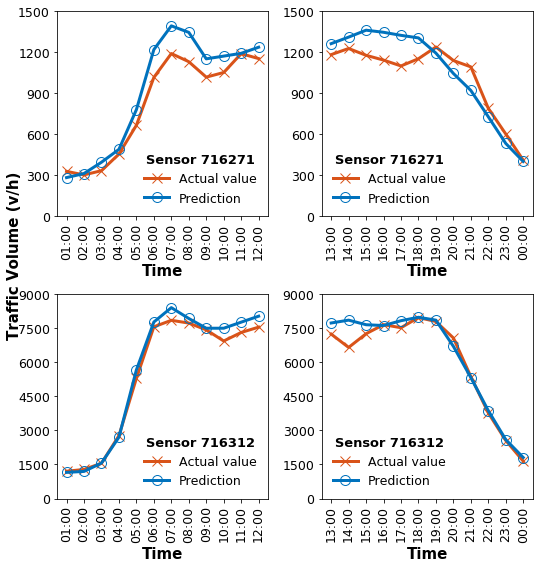}
		\caption{The predictions made by \model{} on Sensor 716271 and 716312. LEFT: 12-step predictions starting at 01:00. RIGHT: 12-step predictions starting at 13:00. The predictions, in general, match the actual values well, especially during the nighttime.}
		\label{fig:visual}
\end{figure}

\subsection{Temporal Prediction Performance} 
In order to study the temporal prediction performance in detail, we present MAE, MAPE, and RMSE of all six models on individual prediction steps.

As shown in Fig.~\ref{fig:temporal}, \revise{using the hourly dataset,} all three deep learning models outperform the VAR and LR models over all prediction steps. \revise{\model{} achieves the best performance across all steps except for the first one in Fig.~\ref{fig:temporal}~RIGHT, where DCRNN achieves lower MAPE (0.079) than \model{} (0.082). This is because our deep learning models are trained to minimize the total MAE across the 12 prediction steps, and a lower MAE may not imply a lower MAPE for each step.}

In Fig.~\ref{fig:temporal}, we can also observe that as the prediction step increases, prediction errors of VAR and LR generally increase. HA shows constant prediction errors since it does not use the past 12-step data, but the averaged historical traffic volume across the same time span of a day, to make predictions. \revise{The error curves of DCRNN and Seq2Seq-RNN drop at later prediction steps. Our model shows a similar behavior but with better stability. It is interesting that the error of multi-step prediction does not necessarily increase monotonically. Some studies have also found this phenomenon without further discussion~\cite{hao2019sequence, muralidhar2019multivariate}. This may be because the prediction accuracy at a particular step is a result of many factors including the number of steps, the predictions at previous steps, and the hour of the day. Especially, predicting traffic in peak hours is known to be harder than off-peak hours~\cite{lin2018quantifying,vlahogianni2006statistical}. As we conduct 12-step prediction at one hour per step, the prediction time span may cover both ``easy-to-predict'' hours and ``difficult-to-predict'' hours, thus affecting the shape of the error curve.}


\revise{Fig.~\ref{fig:15mins} shows the results using the 15-minute dataset. \model{} outperforms the other models except for the first step in Fig.~\ref{fig:15mins}~RIGHT, where DCRNN and Seq2Seq-RNN have slightly lower MAPEs. Again, this may be because the total MAE is being minimized over all steps using the training set, which does not guarantee absolute step-wise performance on the test set. Another observation is that the error curves do not drop as much. We think this is because the 12-step 15-minute traffic volumes fluctuate less than the 12-step hourly traffic volumes, i.e., the former may include only ``easy-to-predict'' hours or ``difficult-to-predict'' hours but not both. Hence, the prediction accuracy at later steps may mainly affected by the number of steps and the predictions at previous steps.}


\revise{To further understand how the deep learning models perform in different hours of a day, Fig.~\ref{fig:dynamics} shows MAE at the 1st, 6th, and 12th prediction steps across 24 hours on weekdays. The average traffic volumes are also shown. As a result, predictions of the early-morning hours (04:00-08:00) have relatively large MAE, comparing to the rest of the day. The three deep learning models perform similarly at the 1st prediction step across all hours as shown in Fig.~\ref{fig:dynamics} TOP. However, our model outperforms DCRNN and Seq2Seq-RNN over the early-morning hours at the 6th and 12th prediction steps as shown in Fig.~\ref{fig:dynamics} MIDDLE and BOTTOM.}

\subsection{Spatial Prediction Performance} 

In order to demonstrate the spatial prediction performance, in Fig.~\ref{fig:close}, we show the histograms of MAE, MAPE, and RMSE of \model{} and DCRNN \revise{using the hourly dataset with all 150 sensors}. \revise{These results serve as the ``macroscopic'' examination of the prediction performance.} In general, \model{} achieves better prediction accuracy. For example, Fig.~\ref{fig:close}~LEFT shows that using \model{} most sensors have MAE between 200 and 400, while using DCRNN most sensors have MAE between 400 and 600. \model{} also results in more sensors to have lower MAPE and RMSE than DCRNN. These results are shown in Fig.~\ref{fig:close}~MIDDLE and RIGHT.

\revise{We additionally sample the ``microscopic'' examination of the prediction performance}. In Fig.~\ref{fig:visual}, we present the prediction results on two randomly selected sensors. For each sensor, two 12-step predictions are shown: one starting at 01:00 and the other starting at 13:00. The corresponding actual values are also shown for comparison. In general, our predictions match the actual values well, especially during the nighttime.


\section{Conclusion and Future Work}
\label{sec:conclusion}

We propose Graph Convolutional Gated Recurrent Neural Network (\model{}) for network-wide, multi-step traffic volume prediction. Our approach does not require the predefinition of the adjacency matrix for graph convolution. Instead, it integrates the DDGF with Gated Recurrent Neural Network, which can be used to capture hidden correlations among traffic sensors. The sequence-to-sequence architecture further explores temporal dependencies for multi-step prediction. We have evaluated \model{} using \revise{two real-world datasets with different temporal resolutions} and contrasted it to two other deep learning models and three statistical benchmark models. \model{} outperforms all other models in nearly all experiments. Furthermore, we have demonstrated the advantages of our approach from both the temporal and spatial perspectives. 

The future directions are abundant. First, we plan to analyze the learned adjacency matrices so that we can better understand the correlations among traffic sensors. This knowledge can serve as a prior for building future prediction models. Second, we plan to apply our method to predict other types of traffic data such as traffic speed and vehicle trajectory, in combination with existing techniques~\cite{Lin2020Attention}. This is promising because of the wide adoption of traffic sensing technologies~\cite{lin2019efficient} such as loop detector~\cite{lin2015modeling,Caltrans}, GPS~\cite{zhu2017prediction,zhu2017trajectory,Li2017CityFlowRecon,Li2018CityEstIter}. \revise{Third, we plan to conduct more experiments in order to understand the shape of the error curve of multi-step prediction.} Fourth, we would like to use the predictions of our approach in testing the navigation of autonomous vehicles~\cite{Shen2021Free,Li2019ADAPS} by constructing and utilizing virtual traffic flows~\cite{Wilkie2015Virtual,Chao2019Survey}. Lastly, we are interested in applying our technique to study traffic states under critical conditions such as the pandemic~\cite{Wang2021Mobility,Lin2021Safety}. 

\textbf{Conflict of Interest Statement:
}
On behalf of all authors, the corresponding author states that there is no conflict of interest.

{\small
	\bibliographystyle{IEEEtran}
	\bibliography{reference}
}


\end{document}